\documentclass[letterpaper, 10 pt, conference]{ieeeconf}  

\IEEEoverridecommandlockouts                              

\title{\LARGE \bf
NLOS Ranging Mitigation with Neural Network Model for UWB Localization
}

\author{Muhammad Shalihan, Ran Liu, and Chau Yuen
\thanks{M. Shalihan, R. Liu, and C. Yuen are with the Engineering product Development Pillar, Singapore University of Technology and Design, 8 Somapah Rd, Singapore 487372. \{ran\_liu, yuenchau\}@sutd.edu.sg.}
}
\usepackage{subfigure}
\usepackage{cite}
\usepackage{caption}
\usepackage{amsmath,amssymb,amsfonts}
\usepackage{algorithmic}
\usepackage{graphicx}
\usepackage[rightcaption]{sidecap}
\usepackage{textgreek}
\usepackage{wrapfig}
\usepackage{textcomp}
\usepackage{xcolor}
\usepackage[export]{adjustbox}
\usepackage{float}
\usepackage{pbox}
\def\BibTeX{{\rm B\kern-.05em{\sc i\kern-.025em b}\kern-.08em
    T\kern-.1667em\lower.7ex\hbox{E}\kern-.125emX}}

\begin{document}

\maketitle
\thispagestyle{empty}
\pagestyle{empty}

\begin{abstract}

Localization of robots is vital for navigation and path planning, such as in cases where a map of the environment is needed. Ultra-Wideband (UWB) for indoor location systems has been gaining popularity over the years with the introduction of low-cost UWB modules providing centimetre-level accuracy. However, in the presence of obstacles in the environment, Non-Line-Of-Sight (NLOS) measurements from the UWB will produce inaccurate results. As low-cost UWB devices do not provide channel information, we propose an approach to decide if a measurement is within Line-Of-Sight (LOS) or not by using some signal strength information provided by low-cost UWB modules through a Neural Network (NN) model. The result of this model is the probability of a ranging measurement being LOS which was used for localization through the Weighted-Least-Square (WLS) method. Our approach improves localization accuracy by 16.93\% on the lobby testing data and 27.97\% on the corridor testing data using the NN model trained with all extracted inputs from the office training data. 
\end{abstract}

\section{INTRODUCTION}

Localization is a crucial step to achieving autonomous movement. The literature shows different techniques, such as the Ultra-Wideband (UWB) localization approaches. For example, the authors in \cite{lidar_nlos} proposed a UWB NLOS identification approach using the LiDAR point cloud. With the LiDAR point cloud map, NLOS ranging measurements can be identified by judging whether obstacle occlusions exist between the mobile tag and anchor. Authors in \cite{monocular} proposed an enhanced tightly coupled sensor fusion scheme using a monocular camera and UWB ranging to perform Simultaneous Localization And Mapping (SLAM). UWB is a robust solution in Global Navigation Satellite System (GNSS) denied environments. It can provide accuracy up to a centimetre level, making it ideal for replacing complex and expensive motion-capture approaches such as with the LiDAR or monocular camera \cite{UWB_multiuav}.

Research work to utilize UWB in various localization applications is shown in the literature. Localization can be achieved with UWB modules installed on quadcopters to actively record ranging information, which can be exchanged between different quadcopters to determine the position of its neighbouring Unmanned Aerial Vehicle (UAV) relative to itself. To account for the lack of information from distance-only measurements for the relative position estimate, a combination of nonlinear and linear trajectory \cite{Quadcopter_uwb} was utilized. 

A similar approach proposes using multiple ground robots and achieves relative localization. By determining the position of the other robots relative to itself, the approach then 

\begin{figure}[H]
\centering
\includegraphics[width=8.7cm, height=5.0cm]{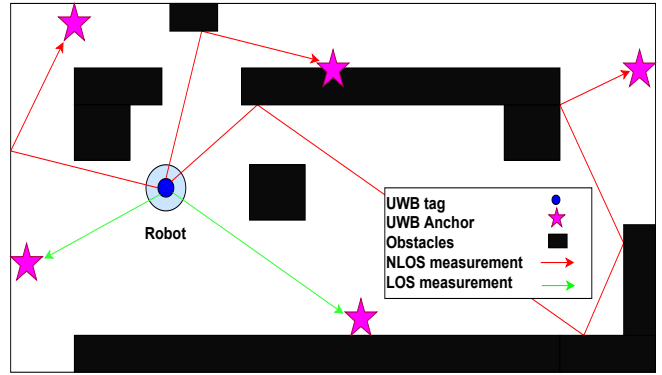}
\caption{Overview of the problem where UWB tag on robot may receive longer ranging measurements due to NLOS conditions. This will affect localization accuracy if not mitigated.}
    \label{fig:mesh1}
\end{figure}

\noindent fuses the optimized pose with odometry information \cite{zhiqiang}. Authors in  \cite{Graph_optimization_UWB} used UWB ranging measurements in a graph optimization approach with an outlier rejection algorithm to mitigate the effects of NLOS signals. However, the outlier rejection was achieved by setting fixed criteria in an inequality equation, which may not generalize well for different NLOS scenarios. While both these works produce significant improvements for relative localization, not dealing with Non-Line-Of-Sight (NLOS) situations effectively may cause the results to degrade significantly.

Recent research shows the popularity of UWB for positioning, especially within the robotics community, due to its low cost, low power consumption, and small size. Robust localization can be achieved through the use of UWB ranging measurements fused with data from onboard sensors such as the Inertial Measurement Unit (IMU) through an Extended Kalman Filter (EKF) and utilizing UWB based communication to transfer the quadcopter's orientation as proposed in \cite{RL_UWB_EKF_IMU}. UWB ranging measurements are also fused with visual odometry to remove visual drift and improve robustness through a pose graph optimization from visual tracking constraints and the proposed smoothness and range constraints \cite{UWB_fastloc_mapping}. Although UWB can provide up to centimetre-level accuracy and a maximum reading range of up to 100 metres, NLOS measurements could significantly degrade localization results. Figure \ref{fig:mesh1} shows a scenario where the robot will receive measurements from UWB anchors which are blocked by obstacles. UWB measurements are based on the signal Time Of Arrival (TOA) or Time Difference Of Arrival (TDOA) \cite{NLOS_lowcostuwb}. Obstacles and occlusions in the environment would cause the TOA to be much larger than it is, leading to a longer ranging measurement. On the other hand, the TDOA could be smaller. This error in ranging measurement due to NLOS conditions causes sub-optimal localization results. Removing these measurements would not be ideal as it may cause significant errors since not enough measurements were considered for localization, making it essential to identify these NLOS measurements and mitigate their effect on localization.

The effort to mitigate NLOS range measurements to improve localization accuracy can be seen in the literature. For example, \cite{Graph_optimization_UWB} proposes an outlier rejection algorithm for NLOS scenarios. This was done by setting criteria in an inequality equation to determine if the range measurement satisfies this equation. The range constraint is rejected and not included in the cost function if the inequality is satisfied. However, excluding all measurements deemed NLOS may result in worse results because there may not be enough range measurements to get an accurate relative estimation of the robot position. Instead, the ranging measurements deemed NLOS should be mitigated and included in localization.

Therefore, this paper proposes an approach to identify NLOS ranging measurements by giving a probability to each ranging measurement used for localization through Weighted-Least-Square (WLS). We first collected data between a UWB tag carried on a moving robot and fixed UWB anchors in the environment in NLOS conditions, followed by Line-Of-Sight (LOS) conditions. This data is used to train a NN model capable of computing the probability that a ranging measurement is LOS. To verify the localization accuracy, the model trained previously is then used in a series of experiments on the data collected in different environments by using the probability of ranging measurement being LOS computed by the NN model in a WLS localization and comparing the results to the Non-Weighted Least-Square (NWLS) localization. The effects of having more NLOS mitigated are also evaluated in the corridor environment. The contributions of this paper are summarized as follows:
\begin{itemize}
\item We built a NN model trained using the data we collected and labelled according to its LOS and NLOS conditions, which returns a probability of the ranging measurement being LOS based on the UWB data provided.
\item We performed WLS localization using the probabilities generated for each ranging measurement through our NN model and compared the results with NWLS localization.
\item We have conducted experiments (in office, lobby and corridor environments) to evaluate the accuracy of the proposed approach and improve the localization accuracy by 16.93\% on the lobby testing data and 27.97\% on the corridor testing data using the NN model trained with all extracted inputs from the office training data.
\end{itemize}
The remainder of this paper is as follows: Section II introduces the related work. Section III introduces the method and design of our approach. Extensive experiments were carried out in Section IV to validate the effectiveness of the proposed approach. Finally, we conclude in Section V.

\section{Related Work}
Research and literature in the robotics field show a growing interest in navigation and path planning localization. The use of the Global Positioning System (GPS) cannot be applied indoors as the signals from satellites are efficiently diffracted by surrounding buildings and obstructions \cite{NLOS_WLS_CS}. This leads to a growing interest in research focused on localization in GPS-denied environments using UWB devices with different methods of mitigating the NLOS errors. For example, authors in \cite{Range_beacon} proposed a robust range-only beacon localization through trilateration and implementing a voting scheme where each consistent measurement pair "votes" for its solution. The most significant number of votes will be taken as the location estimate. However, the method is not robust to NLOS scenarios which could degrade localization results.

Authors in \cite{NLOS_WLS_CS} proposed to model and characterize amplitude and delay statistics of UWB channels and identify NLOS signals using the model. WLS localization is then performed to evaluate their performances. Authors in \cite{NLOS_DL} proposed to employ a method based on Convolutional Neural Network (CNN) to extract non-temporal features from raw channel impulse response (CIR) signals from the UWB and Long-Short Term Memory Recurrent Neural Network (LSTM-RNN) to classify LOS/NLOS signals. Finally, authors in \cite{UWB_Trilateration} proposed a robust trilateration method by using a confidence-based trilateration method using data from the Multipath Fading Channel (MPF). However, data from channel statistics may not be available in many scenarios in low-cost UWB devices. 

Authors in \cite{NLOS_lowcostuwb} proposed an NLOS identification and mitigation technique for low-cost UWB devices using different machine learning algorithms on data that were collected specifically for two situations when in NLOS. Authors in \cite{NLOS_IMU_UWB} proposed an NLOS error compensator using the principle of inertia through a "virtual inertial point" built from the IMU and the motion trend of the UWB. Authors in \cite{NLOS_Sensornode_loc} proposed a New Hypothesis Test method for NLOS identification and mitigation by comparing the Mean Square Error (MSE) of the range estimates with the variance of the LOS range estimates. However, these approaches can replace their NLOS mitigation techniques with a NN model to classify signals better without complex architectures.

\section{Proposed NLOS Identification Method \& Design}
\begin{figure}[H]
\centering
\includegraphics[width=8.8cm, height=5cm]{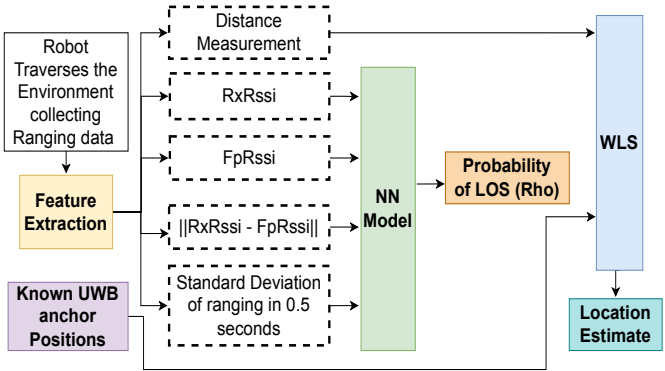}
\caption{Overview of NLOS mitigation localization process.}
    \label{fig:mesh2}
\end{figure}

An overview of the approach is shown in Figure \ref{fig:mesh2}. In this section, we will explain the input data extracted from the UWB tag carried by the moving robot, which consists of the distance measurement, the Residual Received Signal Strength Indicator (RxRssi), the First Path Received Signal Strength Indicator (FpRssi), the difference between Received Signal Strength (RSSD), and the Standard Deviation (STD) of ranging measurements made between a 0.5-second window, between UWB tag carried by the moving robot and fixed UWB anchors in the environment. These input data will be fed 

\noindent into our trained NN model to generate the probability of the ranging measurement being LOS, which refers to a number between 0 to 1, used as the weight for the ranging measurement in the WLS localization.

\subsection{System Overview}
UWB plays an essential role in range-only localization in GPS-denied areas. However, even though UWB is robust to NLOS scenarios due to its ability to penetrate different materials such as walls, metals, and liquids \cite{UWB_adapted_error_map}, NLOS measurements will return a larger ranging measurement, which will affect localization accuracy. Therefore, NLOS measurements need to be identified and mitigated.

Low-cost UWB devices do not provide data from channel statistics used in other NLOS ranging mitigation approaches. Therefore, the NN model in this work can be applied in different environments to mitigate the effects of NLOS ranging measurements, using only data available from low-cost UWB devices. The model returns a probability of the ranging measurement being LOS, based on the UWB data extracted from the UWB node carried by the robot and fed into the NN model. The probability, distance measurement and known UWB anchor locations are then used in a WLS localization process to estimate the robot's position.    
 
\subsection{Neural Network Model}
The NN model used in this paper was built using Tensor Flow, an open-source deep-learning library for building \cite{TensorFlow}. 
UWB data collected from the UWB tag carried by the moving robot in different environments is used as the model's input vector. The input data then goes through the fully connected layers, and finally, the probability of ranging measurement being LOS is received at the output layer.
\subsubsection{Input Features}
The input layer takes in data extracted from the UWB node attached to the robot as it traverses the environment. As shown in Figure \ref{fig:mesh2}, the feature extraction process refers to the use of the RxRssi, which refers to the relative power present in the received signal, and FpRssi, which refers to the power present in this first signal detected, at each timestamp to calculate the RSSD, and using the timestamp and ranging measurement to calculate the STD within a 0.5-second window. Experiments were carried out with the
\begin{figure}[H]
\centering
\includegraphics[width=9.0cm, height=7.0cm,left]{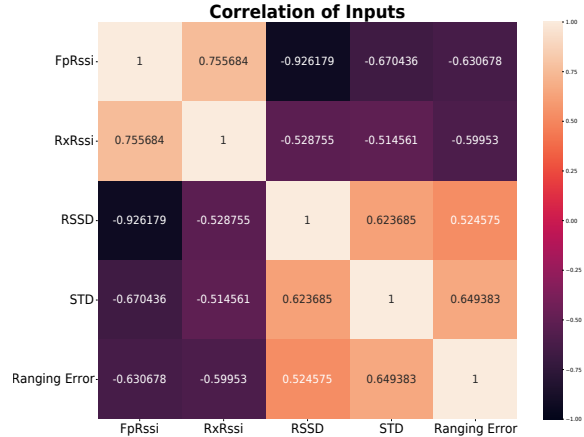}
\caption{Correlation of input data vs ranging error}
    \label{fig:idvre}
\end{figure}

\noindent NN models trained with different combinations of input data. Input data includes the RxRssi and FpRssi extracted from the UWB node at every timestamp, the RSSD, which is also a good indicator for LOS/NLOS conditions, and lastly, the STD of ranging measurements made within a 0.5-second window between the UWB node attached to the robot and each anchor as a higher STD will indicate that there is an NLOS ranging measurement \cite{Survey_NLOS_TOA}. The relationship between the inputs and the ranging error is represented as Spearman's coefficient and expressed in a heatmap in Figure \ref{fig:idvre}. It assesses how well two variables correlate to each other with a monotonic function \cite{corelation}. Spearman's coefficient ranges from -1 to +1, where -1 indicates a perfect negative relationship, 0 indicates no correlation and +1 indicates a perfect positive relationship. A positive relationship indicates that the other variable increases as one variable increases. A negative relationship indicates that the other variable decreases as one variable increases. For example, Figure \ref{fig:idvre} shows that FpRssi and RxRssi have a negative monotonic relation with ranging error.
In contrast, RSSD and STD of ranging have a positive monotonic relationship with ranging error. The combination of inputs chosen through experimentation will go through the Fully Connected layer (FCN), which uses the Rectified Linear Units (ReLU) activation function. Finally, the output layer uses the Sigmoid activation function, which outputs a value between zero to one. This is mainly used in models to predict the probability as an output, which can be used as weights for the localization algorithm.

\subsubsection{Model Architecture}
Different hyperparameters were experimented with during training to select the proper hyperparameters for the NN model. In particular, the number of hidden layers, neurons per hidden layer, learning rate and the number of epochs trained. The results were evaluated based on the model's accuracy and precision on test data. The classification results with different hyperparameters are listed in Table \ref{modelarchitecture} below.

\begin{table}[H]
\caption{Evaluation of hyperparameters for NN model.}
\begin{center}
\begin{tabular}{|c|c|c|c|c|}
\hline
\textbf{Hidden Layers} & \textbf{Neurons}& \textbf{Epochs}& \textbf{Accuracy}& \textbf{Precision} 
\\
\hline
 3 & 100 & 100 & 79.96 & 0.7886 \\ 
\hline
 3 & 300 & 50 & 83.91 & 0.8542\\
\hline
 15 & 100 & 100 & 81.75 & 0.8265\\
\hline
 10 & 300 & 50 & 82.64 & 0.8288\\
\hline
 10 & 300 & 300 & 93.29 & 0.9321\\
\hline

\end{tabular}
\label{modelarchitecture}
\end{center}
\end{table}

Using the best hyperparameters, which produce the best accuracy and precision, as shown in Table \ref{modelarchitecture}, four different models which take different combinations of inputs were trained for NLOS mitigation experimentation.

\subsection{Weighted Least-Square Localization}
The NLOS identification results from the NN model was used to improve the localization results. In this section, we present the WLS localization technique to mitigate the effects of NLOS ranging measurements. By considering the known UWB anchor positions, we can express the WLS estimate of the robot's position as follows \cite{NLOS_WLS_CS}:

\begin{equation}
\hat{x}= \operatorname*{argmin}_{\mathbf{\emph{x}}=(x,y)} \{ \sum_{i=1}^{N} (\beta_i\hat{d_i} -\parallel \mathbf{\emph{x}}-\mathbf{X}_i\parallel)^2 \} \label{eq}
\end{equation}

where \textbeta\textsubscript{\emph{i}},  is the weight for the ranging measurement derived from the probability of a ranging measurement being LOS computed by the NN model, \^{d}\textsubscript{\emph{i}} refers to the ranging measurement to the \emph{i}\textsuperscript{th} anchor, \emph{x} refers to the robot position to be minimized, X\textsubscript{\emph{i}} refers to the location of the \emph{i}\textsuperscript{th} anchor, and \emph{N} is is the number of anchors in the environment. Minimizing \eqref{eq} is done through numerical search methods like Levenberg-Marquardt or Newton's method. In this experiment, Trust Region Reflective (TRF) algorithm was used as it is a generally robust method. To avoid convergence to local minima of the loss function, the constraints and initial guess must be close to the solution \cite{NLOS_WLS_CS}. Therefore, for evaluation purposes, the known starting position of the robot is used for localization, and every estimate for the previous position is used as the initial position for the next estimate.

\section{Experimental Results}

Experimental results presented here are performed on different environments from where the NN model was trained to demonstrate the effectiveness of mitigating NLOS ranging measurements for localization. In addition, results were compared to ground truth obtained with the use of RPLiDAR A3 to perform Adaptive Monte Carlo Localization (AMCL) \cite{amcl} given a map created through GMapping \cite{gmapping}. 

Experiments were carried out using corridor test data with a different number of UWB anchors to test the effects of mitigating more NLOS ranging measurements and including them in localization. 

\subsection{Measurement Campaign}
\subsubsection{Training Data -- Office and Lobby}
In this work, we carried out a measurement campaign for training using low-cost UWB devices in the office and lobby 

\begin{figure}[H]
\centering
\includegraphics[width=7cm, height=6cm]{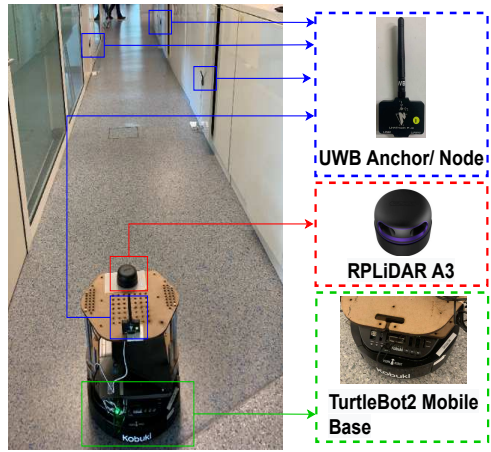}
\caption{Overview of data collection. LOS data was collected along this path, while NLOS data was collected in a path obstructed by the wall.}
    \label{fig:mesh3}
\end{figure}

\noindent of Building 3 at the Singapore University of Technological Design (SUTD), Singapore. 

The training data was collected using LinkTrack UWB devices, a low-cost UWB that uses TDOA protocol that provides IMU information and measures up to 100 metres with a 50Hz sampling rate. As Channel Impulse Response (CIR) must be extracted from the chip, which adds complexity to the system \cite{NLOS_lowcostuwb}, our approach focuses mainly on data readily available from the UWB. Therefore, we only use data recorded by the UWB devices: the timestamp, RxRssi, FpRssi, and distance measurement. Data collected is separated between NLOS and LOS scenarios. For LOS measurements, the robot carrying a UWB node moves along a path with no obstructions collecting UWB data from UWB anchors within LOS as shown in Figure \ref{fig:mesh3}. For NLOS measurements, the UWB anchors were obstructed from the robot during data collection. Data collected were then labelled accordingly at each timestamp, with LOS measurements labelled as 1 and NLOS measurements labelled as 0. 
Each ranging data recorded consists of a timestamp, the distance measurement, RxRssi, FpRssi, the RSSD values, the STD of ranging measurements made within a 0.5-second window, and the LOS/NLOS indicator.

Different inputs were used to train the NN model to determine the inputs that produce the best results.

\subsubsection{Testing Data -- Office, Lobby and Corridor}
Testing data was collected in the office, lobby, and corridor environments. Data collected in this section was used to test the NN model after being trained in a different environment. The robot traversing around the environment carries a single LinkTrack UWB node. Fixed UWB anchors are placed around the environment with known locations. The robot collects ranging measurements, RxRssi, FpRssi and timestamp data from detected UWB anchors as it travels around different environments. At the same time, AMCL poses data is also collected with the help of a map of the environment built through GMapping to be used for ground truth and evaluation purposes.
 
Unlike training data collection, testing data was not separated between LOS or NLOS measurements and therefore contains both LOS and NLOS measurements.
 
\subsection{Lobby Test Data Results Using NN Model Trained Using Office Training Data}

\begin{figure}[H]
\centering
\includegraphics[width=4cm, height=5cm]{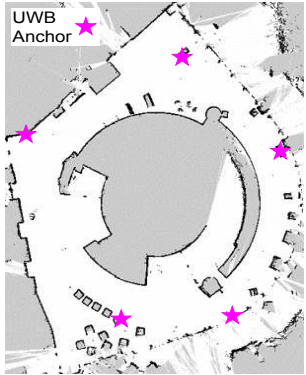}
\caption{Map of lobby environment using GMapping algorithm with 2D-LiDAR.}
    \label{fig:mesh6}
\end{figure} 

The NN model in this section was trained using the office training data mentioned in Section IV.A1 and tested using the lobby testing data mentioned in Section IV.A2. The lobby environment, as shown in Figure \ref{fig:mesh6} is wider as compared to the office environment where the model was trained.

There is a large wall in the lobby environment where the robot traverses. The wall causes much more ranging measurements to be NLOS due to the signals from the UWB anchors being reflected off the walls to reach the UWB node carried by the robot or the fixed UWB anchors placed around the lobby environment.

\begin{figure}
  \centering
  \subfigure[WLS using NN model trained using office training data with all inputs vs NWLS localization tested on lobby testing data using 5 UWB anchors.]
  {\includegraphics[width=8.0cm]{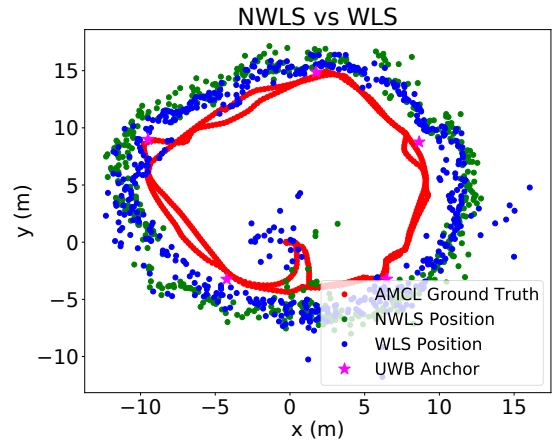}
  \label{fig:lobby1}
  }
  \subfigure[CDF for NN models trained with different combinations of input tested on lobby testing data with 5 UWB anchors.]{\includegraphics[width=8.0cm]{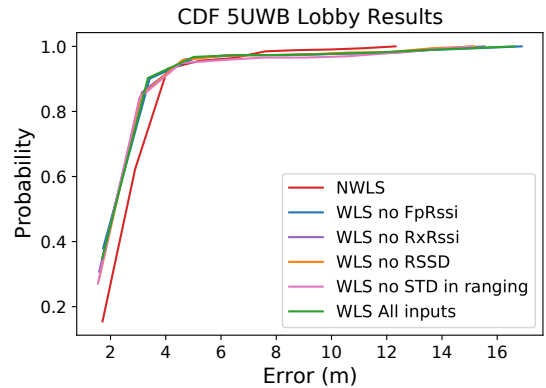} \label{fig:cdf2}}
  \caption{Lobby localization results with 5 UWB anchors} \label{fig:lobbyplots}
\end{figure}

Different combinations of inputs from the office training data were used to train the NN model to determine which inputs produced the best results. The best results according to Table \ref{tab2} and Figure \ref{fig:cdf2} for NWLS vs WLS localization is as shown in Figure \ref{fig:lobby1}. As seen from Table \ref{tab2}, all WLS approaches using different inputs to train the NN model produced improvements in the localization results.

There is a 16.93\% improvement in the estimated position error in terms of metres for WLS localization using the NN
\begin{table}[H]
\caption{Localization results of WLS localization using NN model trained with different combinations of inputs from office training data vs NWLS localization tested on lobby testing data with 5 UWB anchors.}
\begin{center}
\begin{tabular}{|c|c|c|c|c|}
\hline
\textbf{Approach}&\multicolumn{4}{|c|}{\textbf{Error in metres}} \\
\cline{2-5} 
\textbf{Used} & \textbf{\textit{Mean}}& \textbf{\textit{Median}}& \textbf{\textit{SD}}& \textbf{\textit{Improvement}} 
\\
\hline
NWLS& 2.79& 2.65& 1.43& Nil	 \\
\hline
WLS All inputs& 2.32& 2.01& 2.03& 16.93\% \\
\hline
WLS No FpRssi& 2.35& 2.03& 2.05& 16.06\% \\
\hline
WLS No RxRssi& 2.33& 2.04& 2.00& 16.65\% \\
\hline
WLS No RSSD& 2.37& 2.07& 1.92& 15.34\% \\
\hline
WLS No STD& 2.44& 2.06& 2.16& 12.75\% \\
\hline

\end{tabular}
\label{tab2}
\end{center}
\end{table}

\noindent model trained with all inputs compared to NWLS localization using AMCL pose as ground truth. Figure \ref{fig:cdf2} shows the Cumulative Distribution Function (CDF) for WLS and NWLS localization approaches using different models that indicate there is a higher probability of having a minor error with WLS localization as compared to NWLS. The results show that the NN model can mitigate the effects of NLOS ranging measurements and improve localization results for WLS localization compared to NWLS localization.
\subsection{Corridor Test Results Using NN Model Trained Using Office Training Data}
\begin{figure}[H]
\centering
\includegraphics[width=4cm, height=5cm]{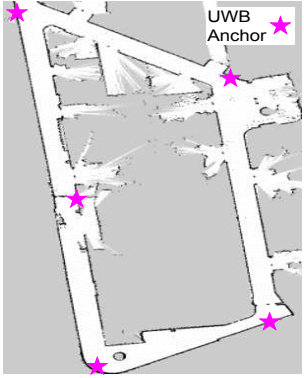}
\caption{Map of corridor environment using GMapping algorithm with 2D-LiDAR.}
\label{fig:mesh8}
\vspace{-0.3cm}
\end{figure}

The NN model in this section was trained using the office training data mentioned in Section IV.A1 and tested using the corridor testing data mentioned in Section IV.A2. The approach was experimented on in a larger map as shown in Figure \ref{fig:mesh8}, which will cause fewer anchors to be detected for localization and more NLOS ranging measurements recorded at each timestamp. Experiments were carried out in this environment using 5, 8 and 9 UWB anchors placed 

\begin{figure}[H]
  \centering
  \subfigure[WLS using NN model trained using office training data with all inputs vs NWLS localization tested on corridor testing data using 5 UWB anchors.]{\includegraphics[width=8.0cm, height=6.0cm]{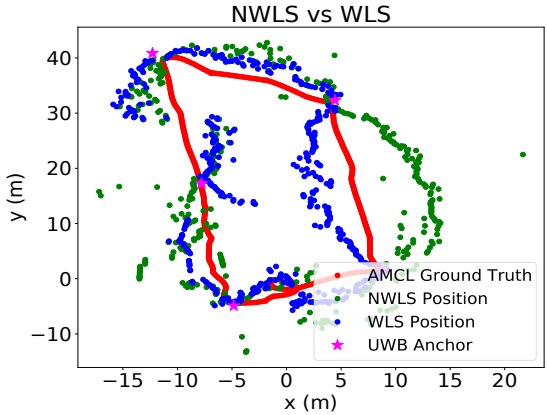}
  \label{fig:corridor1}} \\
      \subfigure[CDF for NN models trained with different combinations of input tested on corridor testing data with 5 UWB anchors.]{\includegraphics[width=8.0cm, height=6.0cm]{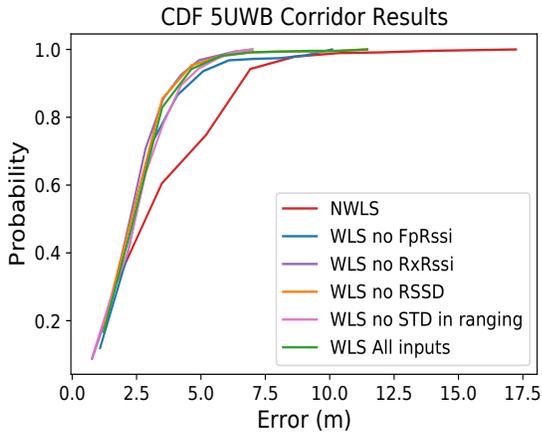} \label{fig:cdf3}}\\
  \caption{Corridor localization results with 5 UWB anchors} \label{fig:corridorplots1}
\end{figure}

\noindent in the environment. For experiments carried out with 5 UWB anchors placed in the environment, as seen from Table \ref{tab3}, all WLS approaches using different combinations of input extracted from the office testing data to train the NN model produced improvements in the localization results. Using the

\begin{table}[H]
\caption{Localization results of WLS localization using NN model trained with different combinations of inputs from office training data vs NWLS localization tested on corridor testing data with 5 UWB anchors.}
\begin{center}
\begin{tabular}{|c|c|c|c|c|}
\hline
\textbf{Approach}&\multicolumn{4}{|c|}{\textbf{Error in metres}} \\
\cline{2-5} 
\textbf{Used} & \textbf{\textit{Mean}}& \textbf{\textit{Median}}& \textbf{\textit{SD}}& \textbf{\textit{Improvement}} 
\\
\hline
NWLS& 3.41& 2.88& 2.47& Nil	 \\
\hline
WLS All inputs& 2.45& 2.36& 1.39& 27.97\% \\
\hline
WLS No FpRssi& 2.65& 2.43& 1.64& 22.09\% \\
\hline
WLS No RxRssi& 2.35& 2.26& 1.23& 31.10\% \\
\hline
WLS No RSSD& 2.36& 2.28& 1.38& 30.57\% \\
\hline
WLS No STD& 2.52& 2.45& 1.36& 25.11\% \\
\hline
\end{tabular}
\label{tab3}
\end{center}
\end{table}

\begin{table}[H]

\caption{Localization results of WLS localization using NN model trained with different combinations of inputs from office training data vs NWLS localization tested on corridor testing data with 8 UWB anchors.}
\begin{center}
\begin{tabular}{|c|c|c|c|c|}
\hline
\textbf{Approach}&\multicolumn{4}{|c|}{\textbf{Error in metres}} \\
\cline{2-5} 
\textbf{Used} & \textbf{\textit{Mean}}& \textbf{\textit{Median}}& \textbf{\textit{SD}}& \textbf{\textit{Improvement}} 
\\
\hline
NWLS& 2.97& 2.64& 1.96& Nil	 \\
\hline
WLS All inputs& 1.72& 1.23& 1.53& 41.98\% \\
\hline
WLS No FpRssi& 1.75& 1.25& 1.45& 41.14\% \\
\hline
WLS No RxRssi& 1.68& 1.21& 1.46& 43.49\% \\
\hline
WLS No RSSD& 1.66& 1.20& 1.43& 44.05\% \\
\hline
WLS No STD& 1.70& 1.24& 1.36& 42.81\% \\
\hline
\end{tabular}
\label{tab4}
\end{center}
\end{table}

\begin{table}[H]
\caption{Localization results of WLS localization using NN model trained with different combinations of inputs from office training data vs NWLS localization tested on corridor testing data with 9 UWB anchors.}
\begin{center}
\begin{tabular}{|c|c|c|c|c|}
\hline
\textbf{Approach}&\multicolumn{4}{|c|}{\textbf{Error in metres}} \\
\cline{2-5} 
\textbf{Used} & \textbf{\textit{Mean}}& \textbf{\textit{Median}}& \textbf{\textit{SD}}& \textbf{\textit{Improvement}} 
\\
\hline
NWLS& 3.39& 3.29& 1.75& Nil\\
\hline
WLS All inputs& 1.42& 1.18& 0.91& 58.02\%\\
\hline
WLS No FpRssi& 1.53& 1.45& 0.92& 54.79\%\\
\hline
WLS No RxRssi& 1.50& 1.41& 0.91& 55.72\%\\
\hline
WLS No RSSD& 1.50& 1.39& 0.92& 55.62\%\\
\hline
WLS No STD& 1.55& 1.45& 0.91& 54.15\%\\
\hline
\end{tabular}
\label{tab5}
\end{center}
\end{table}

\noindent NN model trained with all inputs, a 27.97\% improvement in the estimated position error in terms of metres was achieved for WLS localization compared to NWLS localization. 
The results of the experiment using 5 UWB anchors in the environment according to Table \ref{tab3} is as shown in Figure \ref{fig:cdf3}. Figure \ref{fig:cdf3} shows that the probability of minor error in localization for WLS approaches is higher than that of NWLS localization. Figure \ref{fig:corridor1} shows the localization results using the NN model trained with all inputs compared to NWLS localization. This shows that our approach can improve localization results by using the weights computed by the NN model. Furthermore, our approach includes all NLOS ranging measurements, which was mitigated through a WLS localization as excluding NLOS ranging measurements from localization could produce worse results. For experiments conducted with 8 UWB anchors as seen in Table \ref{tab4} and 9 UWB anchors as seen in Table \ref{tab5}, all WLS approaches using different inputs to train the NN model produced improvements in the localization results. The results for the WLS localization using the model trained with all inputs improved by 41.98\% with 8 UWB anchors and improved by 58.02\% with 9 UWB anchors compared to NWLS localization.

\subsection{Office And Corridor Test Results Using NN Model Trained using Lobby Test Data}
The NN model in this section was trained using the lobby training data mentioned in Section IV.A1 and tested using the lobby and corridor testing data mentioned in Section IV.A2. A model was trained using LOS and NLOS data collected from lobby training data to prove that the approach can improve localization accuracy. The data was collected separately and labelled accordingly, as mentioned in Section IV-A1. The NN model trained with all inputs shown in Table \ref{tab:E2} shows localization accuracy improving by 43.64\% with WLS on the office testing data. There is also a significant improvement on the corridor testing data, as shown in Table \ref{tab:E3}.  

\begin{table}[H]
        \caption{Localization results of WLS localization using NN model trained using lobby training data with all inputs vs NWLS localization tested on office testing data with 5 UWB anchors.}
        \centering
        \begin{tabular}{|c|c|c|c|c|}
        \hline
        \textbf{Approach}&\multicolumn{4}{|c|}{\textbf{Error in metres}} \\
        \cline{2-5} 
        \textbf{Used} & \textbf{\textit{Mean}}& \textbf{\textit{Median}}& \textbf{\textit{SD}}& \textbf{\textit{Improvement \%}} 
        \\
        \hline
        NWLS& 2.02& 1.98& 0.67& Nil\\
        \hline
        WLS & 1.14& 0.95& 0.65& 43.64\%\\
        \hline
        \end{tabular}
        \label{tab:E2}
        \hfill
        \end{table}
        
    \begin{table}[H]   
        \caption{Localization results of WLS localization using NN model trained using lobby training data with all inputs vs NWLS localization tested on corridor testing data with different number of UWB anchors.}
        \centering
        \begin{tabular}{|c|c|c|c|c|}
        \hline
        \textbf{Approach}&\multicolumn{4}{|c|}{\textbf{Error in metres}} \\
        \cline{2-5} 
        \textbf{Used} & \textbf{\textit{Mean}}& \textbf{\textit{Median}}& \textbf{\textit{SD}}& \textbf{\textit{Improvement}} 
        \\
        \hline
        NWLS 5 UWB& 3.41& 2.88& 2.47& Nil\\
        \hline
        WLS 5 UWB& 2.99& 2.75& 2.09& 12.20\%\\
        \hline
        NWLS 8 UWB& 2.97& 2.64& 1.96& Nil\\
        \hline
        WLS 8 UWB& 2.09& 1.53& 1.99& 29.68\%\\
        \hline
        NWLS 9 UWB& 3.39& 3.29& 1.75& Nil\\
        \hline
        WLS 9 UWB& 2.04& 1.77& 1.49& 39.80\%\\
        \hline
        \end{tabular}
        \label{tab:E3}
     \end{table}

\section{Conclusion}
We proposed an approach to mitigate NLOS ranging measurements for UWB-only localization with a NN Model trained and experimented with in different environments. We showed the effectiveness of mitigating NLOS ranging measurements and including them in localization. Our NN models were trained using UWB data extracted from the office and lobby environments. Experiments with the NN model trained using office training data and tested with lobby testing data improved localization results by 16.93\% for the case of 5 UWB anchors. Experiments using corridor testing data showed an improvement of 27.97\% for the case of 5 UWB anchors, 41.98\% for the case of 8 UWB anchors and 58.02\% for the case of 9 UWB anchors. Experiments with the NN model trained using lobby training data and tested on the office and corridor testing data showed significant improvements in localization accuracy. 

\bibliographystyle{IEEEtran}
\bibliography{bib}

\begin{thebibliography}{10}
\providecommand{\url}[1]{#1}
\csname url@samestyle\endcsname
\providecommand{\newblock}{\relax}
\providecommand{\bibinfo}[2]{#2}
\providecommand{\BIBentrySTDinterwordspacing}{\spaceskip=0pt\relax}
\providecommand{\BIBentryALTinterwordstretchfactor}{4}
\providecommand{\BIBentryALTinterwordspacing}{\spaceskip=\fontdimen2\font plus
\BIBentryALTinterwordstretchfactor\fontdimen3\font minus
  \fontdimen4\font\relax}
\providecommand{\BIBforeignlanguage}[2]{{%
\expandafter\ifx\csname l@#1\endcsname\relax
\typeout{** WARNING: IEEEtran.bst: No hyphenation pattern has been}%
\typeout{** loaded for the language `#1'. Using the pattern for}%
\typeout{** the default language instead.}%
\else
\language=\csname l@#1\endcsname
\fi
#2}}
\providecommand{\BIBdecl}{\relax}
\BIBdecl

\bibitem{lidar_nlos}
Z.~Chen, A.~Xu, X.~Sui, C.~Wang, S.~Wang, J.~Gao, and Z.~Shi,
  ``Improved-uwb/lidar-slam tightly coupled positioning system with nlos
  identification using a lidar point cloud in gnss-denied environments,''
  \emph{Remote Sensing}, vol.~14, no.~6, p. 1380, 2022.

\bibitem{monocular}
T.~H. Nguyen, T.-M. Nguyen, and L.~Xie, ``Tightly-coupled ultra-wideband-aided
  monocular visual slam with degenerate anchor configurations,''
  \emph{Autonomous Robots}, vol.~44, no.~8, pp. 1519--1534, 2020.

\bibitem{UWB_multiuav}
W.~Shule, C.~M. Almansa, J.~P. Queralta, Z.~Zou, and T.~Westerlund, ``Uwb-based
  localization for multi-uav systems and collaborative heterogeneous
  multi-robot systems,'' \emph{Procedia Computer Science}, vol. 175, pp.
  357--364, 2020.

\bibitem{Quadcopter_uwb}
K.~Guo, Z.~Qiu, W.~Meng, T.~M. Nguyen, and L.~Xie, ``Relative localization for
  quadcopters using ultrawideband sensors,'' in \emph{Proceedings of
  InternationalMicro Air Vechicle Competition and Conference (IMAV)}, 2016, pp.
  243--248.

\bibitem{zhiqiang}
Z.~Cao, R.~Liu, C.~Yuen, A.~Athukorala, B.~K. Kiat~Ng, M.~Mathanraj, and U.-X.
  Tan, ``Relative localization of mobile robots with multiple ultra-wideband
  ranging measurements,'' in \emph{2021 IEEE/RSJ International Conference on
  Intelligent Robots and Systems (IROS)}, 2021, pp. 5857--5863.

\bibitem{Graph_optimization_UWB}
X.~Fang, C.~Wang, T.-M. Nguyen, and L.~Xie, ``Graph optimization approach to
  range-based localization,'' \emph{IEEE Transactions on Systems, Man, and
  Cybernetics: Systems}, vol.~51, no.~11, pp. 6830--6841, 2021.

\bibitem{RL_UWB_EKF_IMU}
T.-M. Nguyen, A.~Hanif~Zaini, C.~Wang, K.~Guo, and L.~Xie, ``Robust
  target-relative localization with ultra-wideband ranging and communication,''
  in \emph{2018 IEEE International Conference on Robotics and Automation
  (ICRA)}, 2018, pp. 2312--2319.

\bibitem{UWB_fastloc_mapping}
C.~Wang, H.~Zhang, T.-M. Nguyen, and L.~Xie, ``Ultra-wideband aided fast
  localization and mapping system,'' in \emph{2017 IEEE/RSJ International
  Conference on Intelligent Robots and Systems (IROS)}, 2017, pp. 1602--1609.

\bibitem{NLOS_lowcostuwb}
V.~Barral, C.~J. Escudero, J.~A. Garc{\'\i}a-Naya, and R.~Maneiro-Catoira,
  ``Nlos identification and mitigation using low-cost uwb devices,''
  \emph{Sensors}, vol.~19, no.~16, p. 3464, 2019.

\bibitem{NLOS_WLS_CS}
{\.I}.~G{\"u}ven{\c{c}}, C.-C. Chong, F.~Watanabe, and H.~Inamura, ``Nlos
  identification and weighted least-squares localization for uwb systems using
  multipath channel statistics,'' \emph{EURASIP Journal on Advances in Signal
  Processing}, vol. 2008, no.~1, p. 271984, 2007.

\bibitem{Range_beacon}
E.~Olson, J.~J. Leonard, and S.~Teller, ``Robust range-only beacon
  localization,'' \emph{IEEE Journal of Oceanic Engineering}, vol.~31, no.~4,
  pp. 949--958, 2006.

\bibitem{NLOS_DL}
C.~Jiang, J.~Shen, S.~Chen, Y.~Chen, D.~Liu, and Y.~Bo, ``Uwb nlos/los
  classification using deep learning method,'' \emph{IEEE Communications
  Letters}, vol.~24, no.~10, pp. 2226--2230, 2020.

\bibitem{UWB_Trilateration}
J.~Li, X.~Yue, J.~Chen, and F.~Deng, ``A novel robust trilateration method
  applied to ultra-wide bandwidth location systems,'' \emph{Sensors}, vol.~17,
  no.~4, p. 795, 2017.

\bibitem{NLOS_IMU_UWB}
X.~Yang, J.~Wang, D.~Song, B.~Feng, and H.~Ye, ``A novel nlos error
  compensation method based imu for uwb indoor positioning system,'' \emph{IEEE
  Sensors Journal}, vol.~21, no.~9, pp. 11\,203--11\,212, 2021.

\bibitem{NLOS_Sensornode_loc}
G.~Shen, R.~Zetik, H.~Yan, S.~Jovanoska, and R.~S. Thomä, ``Localization of
  active uwb sensor nodes in multipath and nlos environments,'' in
  \emph{Proceedings of the 5th European Conference on Antennas and Propagation
  (EUCAP)}, 2011, pp. 126--130.

\bibitem{UWB_adapted_error_map}
X.~Zhu, J.~Yi, J.~Cheng, and L.~He, ``Adapted error map based mobile robot uwb
  indoor positioning,'' \emph{IEEE Transactions on Instrumentation and
  Measurement}, vol.~69, no.~9, pp. 6336--6350, 2020.

\bibitem{TensorFlow}
J.~V. Dillon, I.~Langmore, D.~Tran, E.~Brevdo, S.~Vasudevan, D.~Moore,
  B.~Patton, A.~Alemi, M.~Hoffman, and R.~A. Saurous, ``Tensorflow
  distributions,'' \emph{arXiv preprint arXiv:1711.10604}, 2017.

\bibitem{Survey_NLOS_TOA}
I.~Guvenc and C.-C. Chong, ``A survey on toa based wireless localization and
  nlos mitigation techniques,'' \emph{IEEE Communications Surveys and
  Tutorials}, vol.~11, no.~3, pp. 107--124, 2009.

\bibitem{corelation}
J.~DeVries, ``About bivariate correlation and linear regression,'' 2007.

\bibitem{amcl}
S.~Thrun, W.~Burgard, and D.~Fox, ``Probalistic robotics,'' \emph{Kybernetes},
  2006.

\bibitem{gmapping}
G.~Grisetti, C.~Stachniss, and W.~Burgard, ``Improved techniques for grid
  mapping with rao-blackwellized particle filters,'' \emph{IEEE Transactions on
  Robotics}, vol.~23, no.~1, pp. 34--46, 2007.

\end{thebibliography}

\end{document}